**Prompting Large Language Models for Clinical Temporal Relation Extraction**

Jianping He, Ms.[1], Laila Rasmy, Ph.D.[1], Haifang Li, Ph.D.[2], Jianfu Li, Ph.D.[2], Zenan Sun, MS.[1], Evan Yu, MS.[1], Degui Zhi, Ph.D.[1], Cui Tao, Ph.D.[2]

[1]McWilliams School of Biomedical Informatics UTHealth at Houston, Houston, TX, USA;
[2]Department of Artificial Intelligence and Informatics, Mayo Clinic, Jacksonville, FL, USA

Corresponding author: Cui Tao, Department of AI and Informatics, Mayo Clinic, Jacksonville, FL, USA; 4500 San Pablo Rd S, Jacksonville, FL 32224
E-mail: Tao.Cui@mayo.edu; Tel: 904-956-3256 Fax: 904-956-3359

**ABSTRACT**

**Objective:** This paper aims to prompt large language models (LLMs) for clinical temporal relation extraction (CTRE) in both few-shot and fully supervised settings.
**Materials and Methods:** This study utilizes four LLMs: Encoder-based GatorTron-Base (345M)/Large (8.9B); Decoder-based LLaMA3-8B/MeLLaMA-13B. We developed full (FFT) and parameter-efficient (PEFT) fine-tuning strategies and evaluated these strategies on the 2012 i2b2 CTRE task. We explored four fine-tuning strategies for GatorTron-Base: (1) Standard Fine-Tuning, (2) Hard-Prompting with Unfrozen LLMs, (3) Soft-Prompting with Frozen LLMs, and (4) Low-Rank Adaptation (LoRA) with Frozen LLMs. For GatorTron-Large, we assessed two PEFT strategies—Soft-Prompting and LoRA with Frozen LLMs—leveraging Quantization techniques. Additionally, LLaMA3-8B and MeLLaMA-13B employed two PEFT strategies: LoRA strategy with Quantization (QLoRA) applied to Frozen LLMs using instruction tuning and standard fine-tuning.
**Results:** Under fully supervised settings, Hard-Prompting with Unfrozen GatorTron-Base achieved the highest F1 score (89.54%), surpassing the SOTA model (85.70%) by 3.74%. Additionally, two variants of QLoRA adapted to GatorTron-Large and Standard Fine-Tuning of GatorTron-Base exceeded the SOTA model by 2.36%, 1.88%, and 0.25%, respectively. Decoder-based models with frozen parameters outperformed their Encoder-based counterparts in this setting; however, the trend reversed in few-shot scenarios.
**Discussions and Conclusions:** This study presented new methods that significantly improved CTRE performance, benefiting downstream tasks reliant on CTRE systems. The findings underscore the importance of selecting appropriate models and fine-tuning strategies based on task requirements and data availability. Future work will explore larger models and broader CTRE applications.
**Word Count:** 3996

## BACKGROUND AND SIGNIFICANCE

In recent years, temporal relation extraction (TRE) has attracted growing interest in clinical research due to its crucial role in understanding disease progression, drug efficacy, and patient outcomes [1]. Clinical Temporal Relation extraction (CTRE) aims to extract the temporal information about the patient's past, current, and future status/conditions that is unavailable in structured Electronic Health Records (EHRs) data [2,3]. This process is essential for reconstructing events within clinical documents, providing a chronological understanding of a patient's medical history [1,4]. Accurate recognition and extraction of temporal relations are crucial in improving diagnostic accuracy and informing treatment strategies [4].

Although current temporal relation extraction in the clinical field has improved in performance, some challenges still need to be tackled to further improve current CTRE systems' performance. First, clinical notes have flexible formatting [5,6]. Therefore, medical providers prefer to use atypical grammatical constructions, including a lot of abbreviations and acronyms in clinical notes [7,8], which makes the annotation very troublesome in CTRE [9]. The CTRE annotation has a lower inter-annotator agreement (IAA) than other clinical annotation tasks, including clinical events and temporal mentions annotation tasks [3]. Additionally, the state-of-the-art (SOTA) methods in CTRE achieved the performance that F-measure is 85.7 on the 2012 i2b2 CTRE dataset [10,11], which is not satisfactory and has bad effects on the downstream applications [1]. Therefore, developing a novel CTRE system that could achieve better and also work in few-shot scenarios is an urgent task in this field.

Prompt tuning in Transformer Encoder-based models has demonstrated strong performance in few-shot scenarios across various NLP tasks [12–16]. Based on these findings, we hypothesize that integrating prompt tuning into the Transformer Encoder-based CTRE system could enhance its performance, particularly in few-shot settings. Peng et al. [17] introduced a prompt tuning-based learning architecture for GatorTron, a clinical Transformer Encoder-based LLM, evaluating it on adverse drug events and social determinants of health (SDoH) datasets. However, they did not explore another parameter-efficient fine-tuning (PEFT) method, namely, Low-Rank Adaptation (LoRA), for GatorTron or evaluate its performance on the 2012 i2b2 CTRE task in both few-shot and fully supervised scenarios. On the other hand, Transformer Decoder-based LLMs have shown great potential in clinical NLP tasks [18–20]. To our knowledge, no studies have leveraged Transformer Decoder-based LLMs, like LLaMA3 or MeLLaMA, for CTRE.

This paper aims to prompt current LLMs for CTRE in both few-shot scenarios and fully supervised settings. We employs four LLMs: Transformer Encoder-based models, Gatortron-Base/Large, pretrained on medical domain corpora [21]; a Transformer Decoder-based model, LLaMA3, pretrained on general domain corpora [22]; and a Transformer Decoder-based model, MeLLaMA, pretrained on medical domain corpora [23]. We selected the GatorTron-Base (345M parameters), GatorTron-Large (8.9B parameters), LLaMA3-8B, and MeLLaMA-13B models to develop and evaluate a range of fine-tuning strategies on the 2012 i2b2 CTRE task. Our exploration encompassed both full fine-tuning (FFT) and parameter-efficient fine-tuning (PEFT) approaches.

Notably, GatorTron-Large, LLaMA3-8B, and MeLLaMA-13B have comparable, large parameter counts within their architectures, making them well-suited for applying Quantized Technique to develop PEFT strategies. In contrast, GatorTron-Base, due to its lightweight parameter size, is more appropriate for FFT approaches. Specifically, we developed and evaluated four fine-tuning strategies for GatorTron-Base: (1) Standard Fine-Tuning without the Use of Prompts, (2) Hard-Prompting with Unfrozen LLMs, (3) Soft-Prompting with Frozen LLMs, and (4) LoRA with Frozen LLMs. For GatorTron-Large, we implemented and assessed two PEFT strategies: (1) Soft-Prompting with Frozen LLMs and (2) LoRA with Frozen LLMs (QLoRA), both utilizing Quantization techniques. Additionally, we developed two distinct PEFT strategies specifically tailored to optimize LLaMA3-8B and MeLLaMA-13B: QLoRA with Frozen LLMs employing instruction tuning and QLoRA with Frozen LLMs utilizing standard fine-tuning.

**RELATED WORKS**

The methodologies adopted in CTRE include rule-based methods, machine learning-based methods, deep learning-based methods, and hybrid methods [1]. The early rule-based systems for CTRE primarily relied on hand-crafted linguistic rules and domain-specific knowledge [24,25]. The machine learning methods used in CTRE consist of conditional random fields (CRF) [26,27], support vector machine (SVM) [28], random forest (RF) [29], and logistic regression (LR) [30]. Deep learning-based methods consist of recurrent neural networks (RNN) [31], convolutional neural networks (CNN) [32], LSTM (long short-term memory) [33], and BERT variants [34]. Deep learning systems or hybrid systems tend to outperform other methods [1].

Most SOTA advancements have utilized Transformer-based architectures, particularly BERT variants, to effectively capture long-range dependencies and contextual meanings within biomedical texts. For instance, Chen et al. [35] combined BERT with a 1-dimensional CNN, achieving an F1 score of 70.85% on the 2012 i2b2 CTRE task. Haq et al. [36] introduced BioBERT with an accuracy and speed optimized method for CTRE. This approach has been shown to outperform the basic BERT model on the 2012 i2b2 CTRE task, achieving a macro F1 score of 73.60%. Knez and Žitnik [10] integrated the Entity BERT language model and a Graph Convolutional Network, incorporating common knowledge about events into the CTRE process. This approach demonstrated strong performance, achieving a micro F1 score of 82.04%. Uma et al. [11] introduced a novel architecture that utilizes a Transformer-based Graph Neural Network, integrating textual data with event graph embeddings to predict CTRE between events. This approach achieved an F1 score of 85.70%.

Recently, LLMs have shown exceptional performance across a diverse range of NLP tasks with their ability to understand and generate human-like text. In the biomedical field, for example, GatorTron, LLaMA3, and MeLLaMA have made notable strides in medical concept and relation extraction where they excel at identifying, classifying, and normalizing clinical entities and relations from unstructured healthcare data [17,23,37]. However, despite the significant potential of these models in various clinical NLP tasks, to the best of our knowledge, there is currently no research that has specifically applied these LLMs to CTRE.

In this project, we employed four LLMs: Gatortron-Base(345M)/Large(8.9B) [21], LLaMA3-8B [22], MeLLaMA-13B [23], and developed various fine tuning methods for these four models and evaluated their performance on the 2012 i2b2 CTRE task [38].

**MATERIALS AND METHODS**

**Dataset**

In this study, we employed the 2012 i2b2 Challenge dataset on Temporal Relations in Clinical Narratives [38], referred to as the i2b2 temporal corpus, to assess the performance of our proposed models. The dataset consists of 310 anonymized patient discharge summaries, annotated with three temporal relationships: AFTER, BEFORE, and OVERLAP. Table 1 provides a summary of the i2b2 temporal relation corpus. Further details can be found in the Supplementary File, under the "Dataset" section. We adopted the same preprocessing methodology as outlined in the previous work by Chen et al. [35]. The data preprocessing steps are detailed in the supplementary file, under the "Data Preprocessing" section.

|  | Training Set | Test Set |
| --- | --- | --- |
| Discharge summaries | 190 | 120 |
| Events | 16,468 | 13,594 |
| Temporal expressions | 2,366 | 1,820 |
| Temporal relations | 33,543 | 27,736 |
| BEFORE | 17,513 | 15,113 |
| AFTER | 3,207 | 2,729 |
| OVERLAP | 12,823 | 9,894 |

Table 1. Summary of the i2b2 temporal relation corpus.

**Employed Large Language Models (LLMs)**

This study employs four LLMs: Gatortron-Base (345M)/Large (8.9B) [21]; LLaMA3-8B [22]; and MeLLaMA-13B [23]. Experimental results demonstrate that GatorTron outperformed existing biomedical and clinical Transformers across five tasks and six benchmark datasets [21]. Notably, the LLaMA3 model has outperformed other models of similar sizes [22]. Me-LLaMA, as one of the most comprehensive open-source medical foundation LLMs trained on both biomedical and clinical data, exhibited superior results across diverse general and medical tasks, positioning it as a highly promising choice for medical AI applications [23]. Consequently, this study selected these models for further analysis. To explore PEFT strategies for optimizing LLMs in resource-constrained scenarios—such as data scarcity, and limited computational resources—we need to explore the LLMs versions with smallest parameters: specifically, LLaMA3-8B and

MeLLaMA-13B. To ensure a fair comparison between models with similar large parameter sizes, GatorTron-Large (8.9B) was selected for this study due to its comparable parameter count to LLaMA3-8B and MeLLaMA-13B. Additionally, to thoroughly assess FFT strategies on smaller models, the smallest version of GatorTron (GatorTron-Base (345M)) was also included in the study. The study, therefore, includes GatorTron-Base (345M) and GatorTron-Large (8.9B), LLaMA3-8B and LLaMA3-8B-Instruct, as well as MeLLaMA-13B and MeLLaMA-13B-chat variants [21–23]. Comprehensive background information of these models can be found in the supplementary file, under the "Employed Large Language Models (LLMs)" section.

Figure 1 illustrates the architecture of various LLMs, with Table 2 highlighting key architectural distinctions. GatorTron employs a Transformer Encoder-only design, processing raw text by tokenizing it into IDs using a 50,176-token vocabulary. These IDs pass through an embedding layer with a maximum sequence length of 512, producing embeddings with a 1,024 feature dimension in the Base version and 3,584 in the Large version. These embeddings then go through 24 (Base) or 56 (Large) Transformer Encoder layers, which generate contextualized representations by capturing inter-token relationships. The encoder output is directed to a classification head for the final prediction. Conversely, LLaMA3-8B and MeLLaMA-13B use a Transformer Decoder-only structure. Both models tokenize input text into IDs with vocabularies of 128,256 (LLaMA3-8B) and 32,000 (MeLLaMA-13B). These IDs are embedded into vectors with sequence lengths of 8,192 and 4,096, and feature dimensions of 4,096 and 5,120, respectively. The embeddings proceed through 32 (LLaMA3-8B) or 40 (MeLLaMA-13B) Decoder layers, refining token contexts. A Linear layer and Softmax function generate the final output token, enabling these models to handle various language processing tasks. Figure 2 illustrates the principle of six primary fine-tuning strategies for LLMs.

|  | Model Architecture (layers) | Vocabulary Size | Max Sequence Length | Feature Dimension |
| --- | --- | --- | --- | --- |
| GatorTron-Base | Encoder (24) | 50,176 | 512 | 1,024 |
| GatorTron-Large | Encoder (56) | 50,176 | 512 | 3,584 |
| LLaMA3-8B | Decocer (32) | 128,256 | 8,192 | 4,096 |
| MeLLaMA-13B | Decocer (40) | 32,000 | 4,096 | 5,120 |

Table 2. The primary architectural differences among different LLMs.

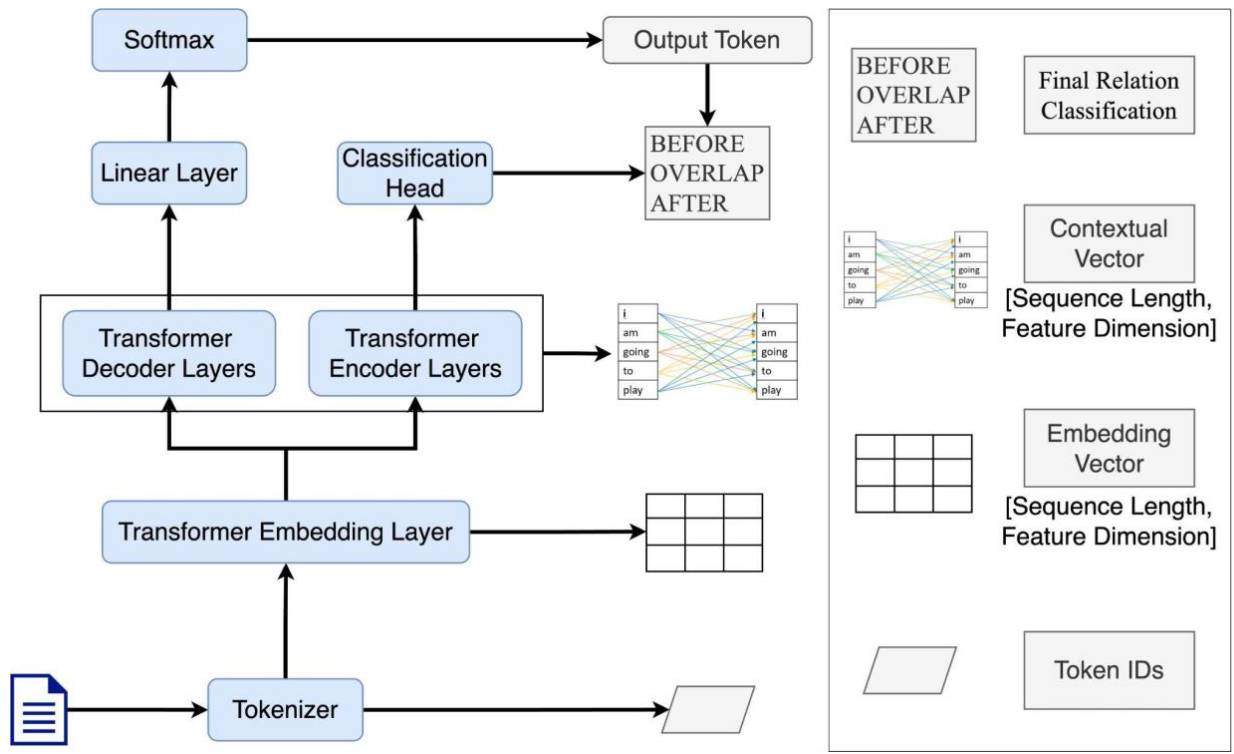

Figure 1. The architectures of LLMs.

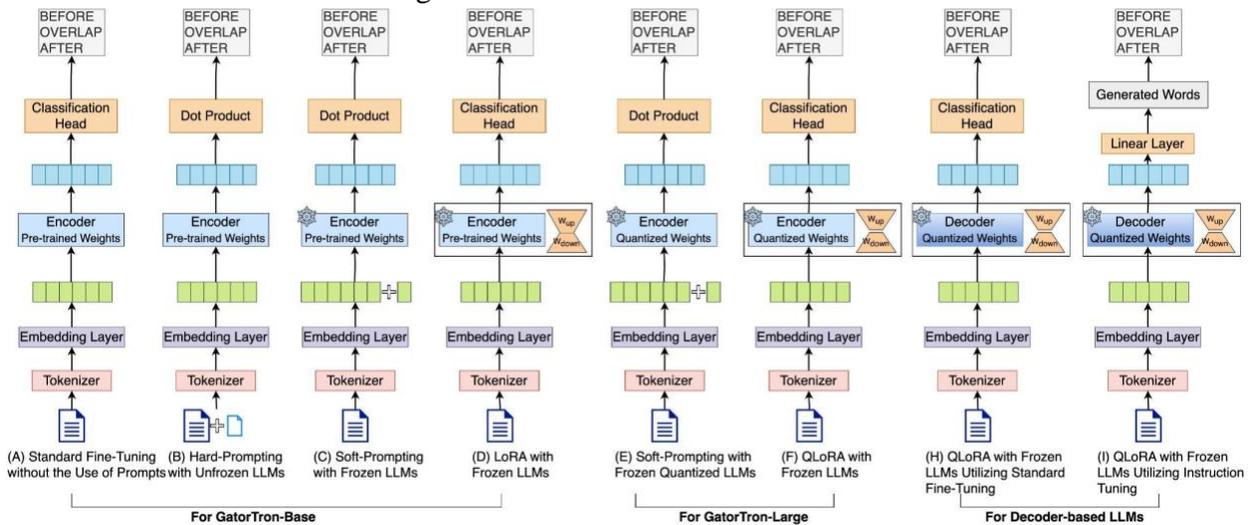

Figure 2. The principle of primary fine-tuning strategies for LLMs.

**Fine-Tuning Strategies for GatorTron-Base**

We developed and evaluated four fine-tuning strategies for GatorTron-Base: (1) Standard Fine-Tuning without the Use of Prompts, (2) Hard-Prompting with Unfrozen LLMs, (3) Soft-Prompting with Frozen LLMs, and (4) LoRA with Frozen LLMs. Figure 3 illustrates the principle of four primary fine-tuning strategies for GatorTron-Base.

(1) Standard Fine-Tuning without the Use of Prompts

In this approach, the model was fine-tuned directly on the CTRE task without using prompts. Subject and object entities in the input text were explicitly tagged with special tokens to emphasize their roles. The input text was tokenized into IDs, processed through a Transformer embedding layer, and passed through Transformer Encoder layers. The resulting contextualized token vectors were then directed to a classification head, yielding the final TREs. In standard fine-tuning, the model remains unfrozen, allowing all layers to be updated throughout the entire fine-tuning process.

(2) Hard-Prompting with Unfrozen LLMs

For hard-prompting, we also kept the LLM weights unfrozen, allowing the model's parameters to be adjusted during training to optimize task performance. This approach employs a structured prompt template, formatted as "Subject [MASK] [MASK] Object," appended to the original input text. By integrating this template, the CTRE task is reformulated into a Masked Language Model (MLM) problem. The model is provided with predefined label words corresponding to each temporal relation class, such as "BEFORE" (labeled as "happened before"), "OVERLAP" (labeled as "happened overlap"), and "AFTER" (labeled as "happened after"). Rather than directly predicting the temporal relation, the model is used to predict the most probable label words to fill the [MASK] positions. The final predicted label words are then mapped to the corresponding temporal relation classification.

Specifically, the contextual vector is first derived by processing the original input text along with the appended hard-prompt template. This process yields the contextual representation of the [MASK] positions by selecting the corresponding slice from the overall contextual vector. Simultaneously, embedding vectors are derived for the predefined label words. The final classification is determined through a dot-product operation between the contextual vectors of the masked tokens and the embeddings of the predefined label words. This operation generates logits for each label word, allowing for the selection of the label word with the highest probability. The selected label word is then mapped to the corresponding TRE.

(3) Soft-Prompting with Frozen LLMs

For soft-prompting, the weights of the language model were kept frozen, while randomly initialized embedding vectors were introduced as soft prompts. These soft prompts were combined with the embedding vectors of the input token sequence. After passing through the Transformer Encoder layers, the soft prompt embeddings were fine-tuned to capture the contextual relationships between entities. Simultaneously, the embedding vectors for the TREs were obtained through tokenization and the Transformer embedding layer. The final TRE was determined through a dot-product operation between the tuned soft-prompts and the embeddings of the classification tokens.

(4) Low-Rank Adaptation (LoRA) with Frozen LLMs

LoRA is a technique that freezes the pretrained model weights while introducing trainable rank decomposition matrices into each layer of the Transformer architecture, significantly reducing the number of trainable parameters required for downstream tasks. For instance, compared to fine-tuning GPT-3 175B using the Adam optimizer, LoRA can decrease the number of trainable parameters by 1,000 times and reduce GPU memory requirements by threefold.

Despite having fewer trainable parameters, LoRA achieves comparable or superior performance to standard fine-tuning [39].

In this study, we also investigated the LoRA method to reduce trainable parameters, enabling the model to adapt to the CTRE task with minimal alteration to the pretrained language model (PLM). LoRA adaptation resembles standard fine-tuning but applies low-rank decomposition to weight matrices in selected Transformer Encoder layers, allowing fine-tuning with reduced computational overhead while keeping most pretrained weights frozen. Token sequence embeddings are processed by LoRA-adapted Encoder layers to generate contextual vectors, which are then passed to a classification head for final TREs.

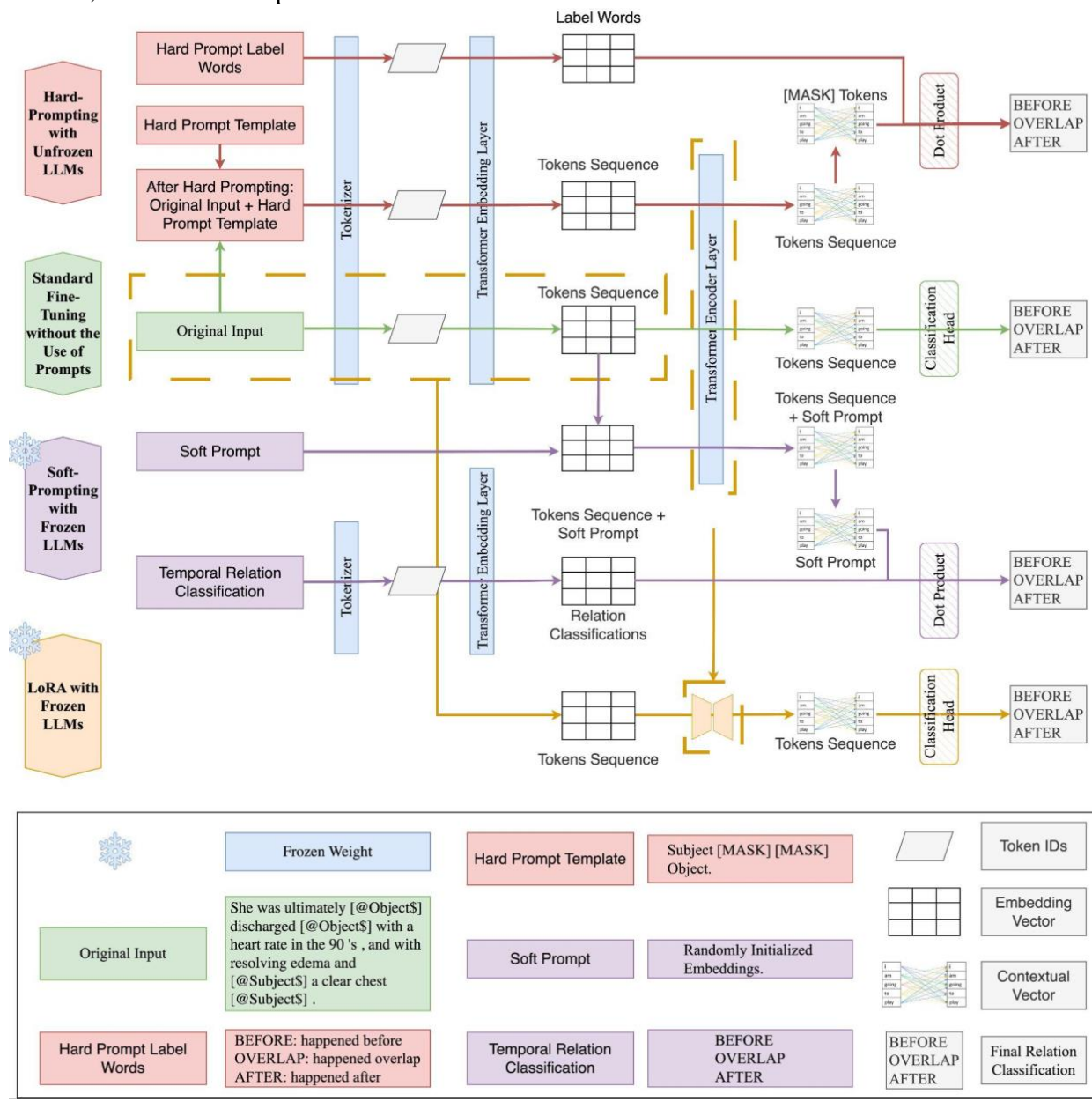

Figure 3. The principle of four primary fine-tuning strategies for GatorTron-Base.

**Parameter-Efficient Fine-Tuning (PEFT) Strategies for GatorTron-Large**

We developed and evaluated two PEFT strategies for GatorTron-Large: (1) Soft-Prompting with Frozen LLMs and (2) LoRA with Frozen LLMs (QLoRA with Frozen LLMs), both implemented using Quantization techniques. These strategies are conceptually similar to their counterparts for GatorTron-Base, with the key distinction being the use of Quantization for GatorTron-Large. Quantization reduces memory and computational demands by representing model weights and activations with lower-precision data types, such as 8-bit integers (int8), enabling efficient deployment of larger models and faster inference [40]. The LoRA strategy with quantization is referred to as QLoRA. In QLoRA, LoRA weights are incorporated into specific layers of the quantized model [41].

**Parameter-Efficient Fine-Tuning (PEFT) Strategies for Transformer Decoder-based Models**

The LLaMA3-Instruct models were developed through instruction tuning, while the LLaMA3 models underwent standard fine-tuning (with a classification head). Our previous study [42,43] demonstrated that reformatting downstream tasks to align with a model's pretraining format significantly enhances performance. To fully utilize these models' capabilities, we applied instruction tuning to the LLaMA3-8B-Instruct model and standard fine-tuning to the LLaMA3-8B model. Similarly, Me-LLaMA base models were obtained through continuous pretraining, with the MeLLaMA-13B-chat model refined via instruction tuning and the MeLLaMA-13B model through standard fine-tuning. Consistent with the approach for LLaMA3 models, additional instruction tuning was performed on the MeLLaMA-13B-chat model and further standard fine-tuning on the MeLLaMA-13B model.

For Transformer Decoder-based models, we applied QLoRA to achieve a balance between performance and efficiency. Specifically, we developed two distinct PEFT strategies to optimize these models for the CTRE task: QLoRA with Frozen LLMs utilizing instruction tuning and QLoRA with Frozen LLMs employing standard fine-tuning. Figure 4 illustrates the principles of these strategies, both of which incorporate prompt engineering to enhance the model's ability to comprehend and analyze temporal relationships effectively.

(1) QLoRA with Frozen LLMs Utilizing Instruction Tuning

The first strategy, QLoRA with frozen LLMs utilizing instruction tuning, employs prompts to guide the model in generating specific temporal relationship terms, such as "BEFORE," "AFTER," or "OVERLAP." The prompt is structured as: *{'role': 'system', 'content': System Message}, {'role': 'user', 'content': Original Input + 'The temporal relationship between the subject and object is?'}, {'role': 'assistant', 'content': [ ]}*. The system message provides detailed instructions (Figure 4). This prompt is designed to direct the model towards identifying and returning the appropriate temporal relationship. After prompt engineering, the Transformer's tokenizer converts the input sequence into token IDs, which are then processed through the embedding and decoder layers to capture the contextual vector of the tokens sequence. Based on the obtained contextual vectors of the token sequence, a linear layer, projecting the 4,096-

dimensional inputs into 128K dimensions, is trained for next-token prediction to generate labels such as "BEFORE," "AFTER," or "OVERLAP."

(2) QLoRA with Frozen LLMs Utilizing Standard Fine-Tuning

The second PEFT strategy, QLoRA with frozen LLMs utilizing standard fine-tuning, Instead of generating a textual label, this method prompts the model to select a temporal classification from a predefined set of numeric values (e.g., {0, 1, 2}), where each number corresponds to a specific temporal relationship. This approach retains a structured prompt but directs the model to output a numeric value that represents the temporal relationship. The structured prompt combines a system message with detailed instructions and the input text, formatted as: *System Message + 'Text: ' + Original Input + 'The temporal relationship between the subject and object is? <Answer>*. While the initial processing follows the same steps as the instruction tuning strategy, this approach differs in the final layer. Instead of a linear layer for next-token prediction, it employs a classification head that projects the 4,096-dimensional input to a 3-dimensional output, directly producing the TRE.

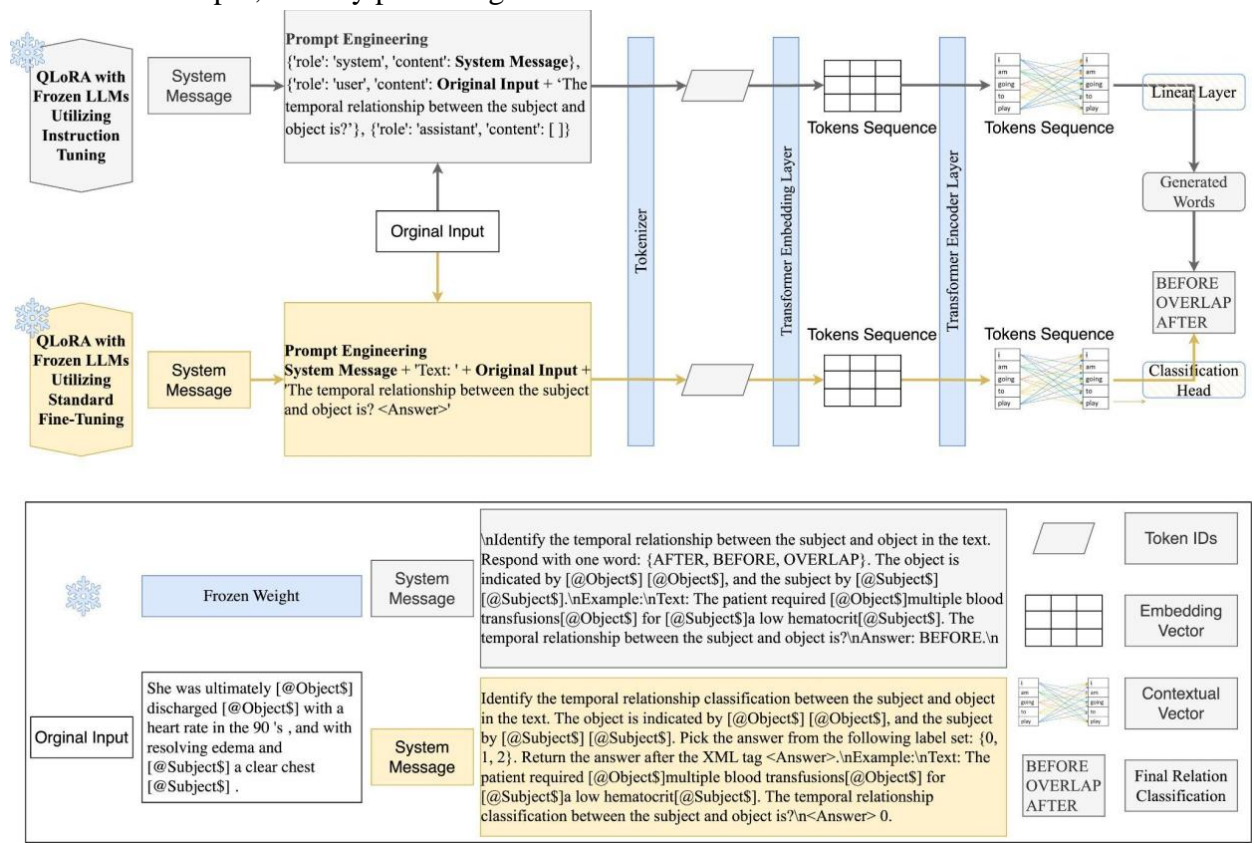

Figure 4. The principle of two primary PEFT strategies for the Transformer Decoder-based models.

Detailed experimental settings for both Transformer Encoder-based models and Transformer Decoder-based models can be found in the Supplementary File under the "Experimental Settings" section.

**RESULTS**

Table 3 presents the model variants evaluated for the CTRE task in this study, including their abbreviations and corresponding full names.

**Fully Supervised Setting**

The SOTA methods for CTRE have achieved an F-measure of 85.70% on the 2012 i2b2 CTRE dataset [10,11]. Table 4 summarizes the performance of various models under the fully supervised setting, measured by Micro F1 scores. Figure 5 visualizes these performances, ranking the models from lowest to highest based on their Micro F1 scores. Notably, the SOTA model is highlighted in grey to emphasize its position among the top-performing models. Among the evaluated models, *GatorTron-Base + hard-prompting* achieved the highest score of 89.54%, surpassing the SOTA model's score of 85.70% by 3.74%. Additionally, *GatorTron-Large* + LoRA$^+$*, *GatorTron-Large* + LoRA$^{++}$*, and *GatorTron-Base* achieved scores of 88.06%, 87.58%, and 85.95%, outperforming the SOTA model by 2.36%, 1.88%, and 0.25%, respectively.

Variants of Transformer Encoder-based models with frozen parameters generally demonstrated lower performance. Specifically, *GatorTron-Base* + LoRA$^+$*, *GatorTron-Base* + LoRA$^{++}$*, *GatorTron-Base* + soft-prompting_32*, *GatorTron-Base* + soft-prompting_64*, *GatorTron-Large* + soft-prompting_32*, and *GatorTron-Large* + soft-prompting_64* achieved relatively low scores of 63.00%, 54.47%, 71.53%, 71.67%, 61.36%, and 69.76%, respectively.

In contrast, Transformer Decoder-based models with frozen parameters generally outperformed their Encoder-based counterparts. In addition, LLaMA3-8B Instruct, fine-tuned using instruction tuning, outperformed LLaMA3-8B fine-tuned using standard methods. Furthermore, LLaMA3-8B Instruct also surpassed all variants of MeLLaMA-13B, regardless of whether they were tuned via instruction tuning or standard fine-tuning. The *LLaMA3-8B-Instruct$^+$* and *LLaMA3-8B-Instruct$^{++}$* achieved scores of 84.50% and 85.00%, respectively, which are only slightly lower than the SOTA model.

Table 3. Model abbreviations and their corresponding full names.

| Model Abbreviations | Corresponding Model Full Names |
|---|---|
| GatorTron-Base | Standard fine-tuning without the use of prompts |
| GatorTron-Base + hard-prompting | Hard-prompting with unfrozen model |
| GatorTron-Base* + soft-prompting_32 | Soft prompting with frozen model utilizing a soft prompt length of 32 |
| GatorTron-Base* + soft-prompting_64 | Soft prompting with frozen model utilizing a soft prompt length of 64 |
| GatorTron-Base* + LoRA+ | LoRA applied to the query and value layers of the Transformer attention mechanism, with the model kept frozen |
| GatorTron-Base* + LoRA++ | LoRA applied to the all linear layers, with the model kept frozen |
| GatorTron-Large* + soft-prompting_32 | Soft prompting with frozen model utilizing a soft prompt length of 32 |
| GatorTron-Large* + soft-prompting_64 | Soft prompting with frozen model utilizing a soft prompt length of 64 |
| GatorTron-Large* + LoRA+ | LoRA applied to the query and value layers of the Transformer attention mechanism, with the model kept frozen |
| GatorTron-Large* + LoRA++ | LoRA applied to the all linear layers, with the model kept frozen |
| LLaMA3-8B-Instruct+ | Instruction tuning applied exclusively to the query and value layers of the Transformer attention mechanism in the model |
| LLaMA3-8B-Instruct++ | Instruction tuning applied to all linear layers in the model |
| LLaMA3-8B+ | Standard fine-tuning with a classification head applied exclusively to the query and value layers of the Transformer attention mechanism in the model |
| LLaMA3-8B++ | Standard Fine-tuning with a classification head applied to all linear layers in the model |
| MeLLaMA-13B-Chat+ | Instruction tuning applied exclusively to the query and |

| Model Abbreviations | Corresponding Model Full Names |
|---|---|
|  | value layers of the Transformer attention mechanism in the model |
| MeLLaMA-13B-Chat[++] | Instruction tuning applied to all linear layers in the model |
| MeLLaMA-13B[+] | Standard fine-tuning with a classification head applied exclusively to the query and value layers of the Transformer attention mechanism in the model |
| MeLLaMA-13B[++] | Standard Fine-tuning with a classification head applied to all linear layers in the model |

Table 4. The performance (micro F1) of various models under the fully supervised setting. *refers to the parameters inside the model kept frozen; [+] denotes LORA applied exclusively to the query and value layers of the Transformer attention mechanism; [++] indicates that LORA adapted to all linear layers.

| Models | Micro F1(%) |
| --- | --- |
| SOTA [11] | 85.70% |
| **GatorTron-Base** | **85.95%** |
| **GatorTron-Base + hard prompting** | **89.54%** |
| GatorTron-Base* + soft-prompting_32 | 71.53% |
| GatorTron-Base* + soft-prompting_64 | 71.67% |
| GatorTron-Base* + LoRA[+] | 63.00% |
| GatorTron-Base* + LoRA[++] | 54.47% |
| GatorTron-Large* + soft-prompting_32 | 61.36% |
| GatorTron-Large* + soft-prompting_64 | 69.76% |
| **GatorTron-Large* + LoRA[+]** | **88.06%** |
| **GatorTron-Large* + LoRA[++]** | **87.58%** |
| LLaMA3-8B-Instruct[+] | 84.50% |
| LLaMA3-8B-Instruct[++] | 85.00% |
| LLaMA3-8B[+] | 77.51% |
| LLaMA3-8B[++] | 69.82% |
| MeLLaMa-13B-chat[+] | 73.15% |
| MeLLaMa-13B[+] | 78.33% |
| MeLLaMa-13B[++] | 72.07% |

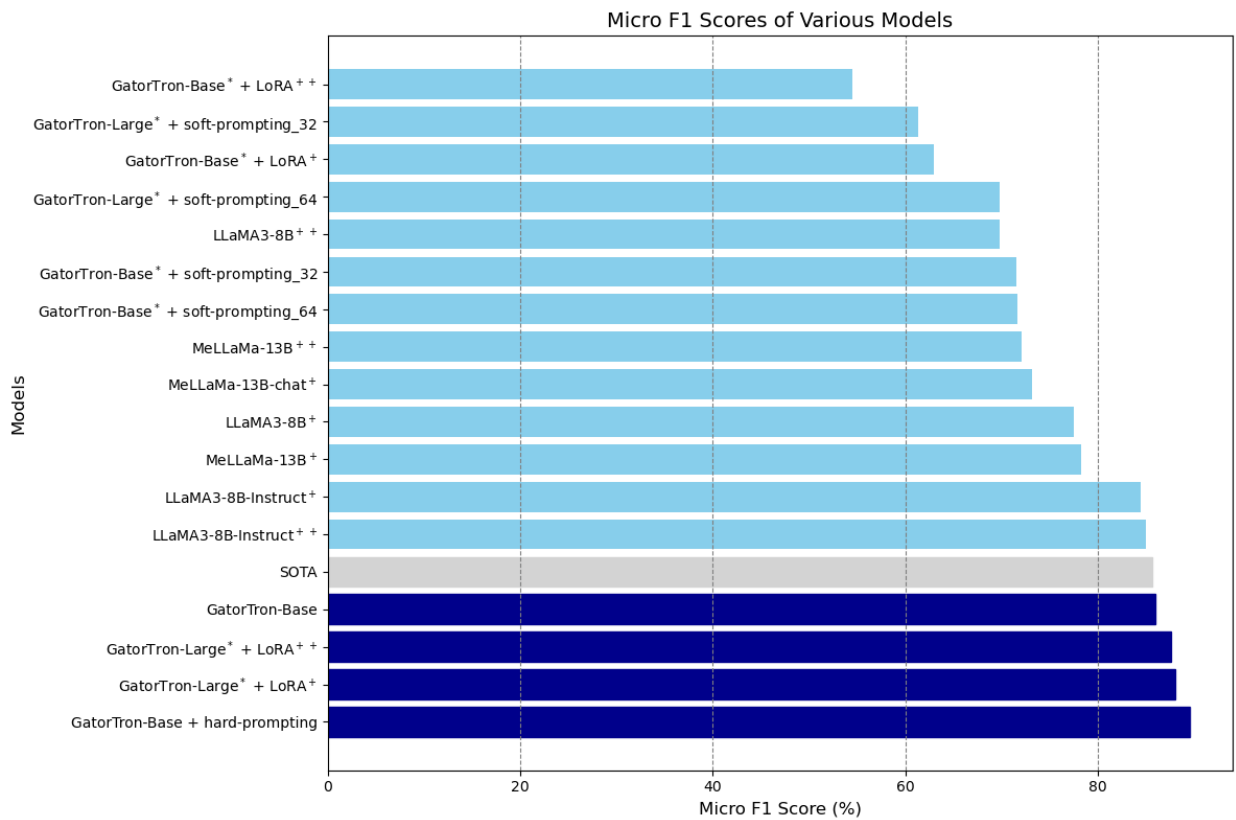

Figure 5. The performance (micro F1) of various models under the fully supervised setting. *refers to the parameters inside the model kept frozen; + denotes LORA applied exclusively to the query and value layers of the Transformer attention mechanism; ++ indicates that LORA adapted to all linear layers.

**Few-Shot Scenarios**

Figure 6 illustrates the performance of various models across different few-shot scenarios, while detailed Micro F1 scores for these scenarios are provided in Table 5 of the Supplementary File. The results indicate a performance increase from the 1-shot to 4-shot settings, with no significant improvement observed beyond the 4-shot setting, including the 8-shot and 16-shot configurations.

Each subplot represents a specific few-shot scenario, with the top half showcasing Transformer Decoder-based models and the bottom half presenting Transformer Encoder-based models. Among the Encoder-based models, the bottom two entries correspond to FFT variants of GatorTron-Base, while the remaining models in the bottom half are Transformer Encoder-based variants with frozen parameters.

In contrast to the trends observed in the fully supervised setting, Transformer Encoder-based models with frozen parameters generally outperformed Transformer Decoder-based models with frozen parameters in few-shot scenarios. Additionally, within the Transformer Decoder-based variants, standard fine-tuning consistently achieved better performance than instruction-tuning approaches.

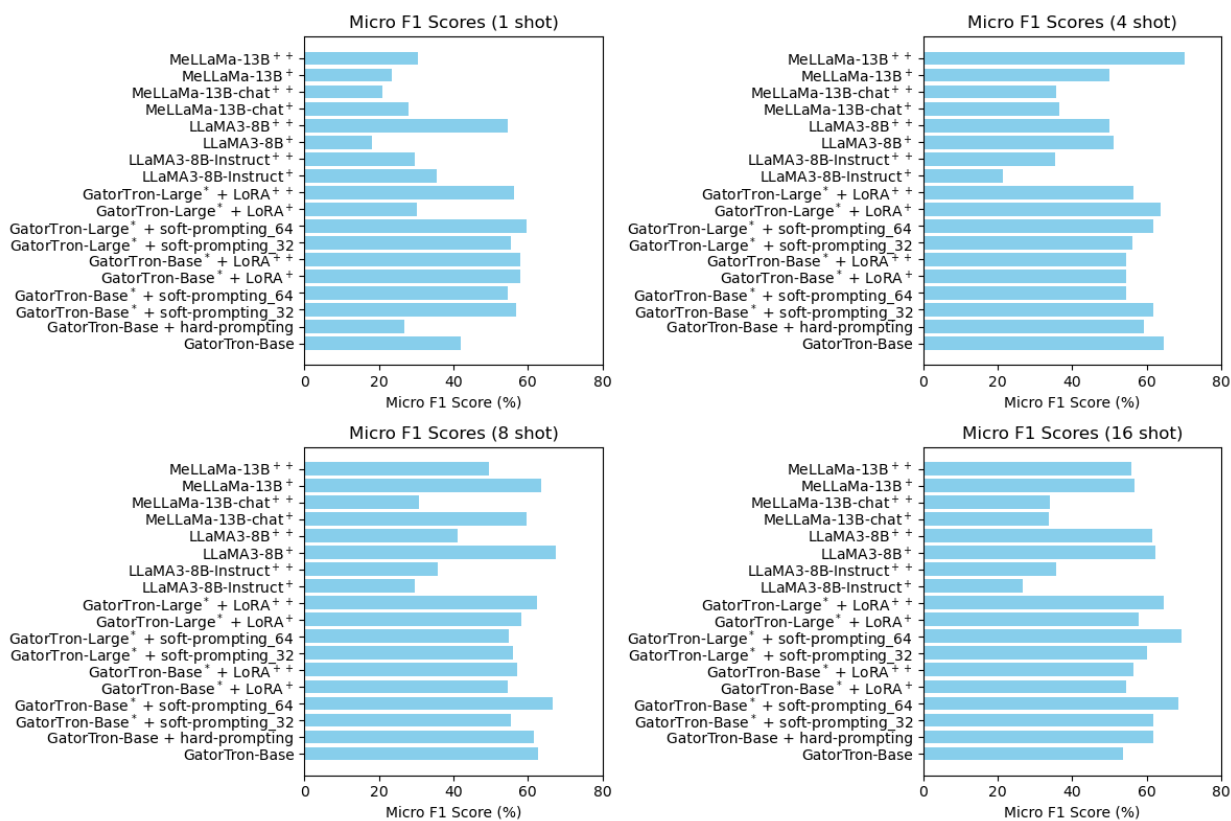

Figure 6. The performance (micro F1) of various models across different few-shot scenarios. *refers to the parameters inside the model kept frozen; + denotes LORA applied exclusively to the query and value layers of the Transformer attention mechanism; ++ indicates that LORA adapted to all linear layers.

# DISCUSSIONS AND CONCLUSIONS

Under the fully supervised setting, the proposed fine-tuning strategies—"*GatorTron-Base + hard-prompting*", "*GatorTron-Large\* + LoRA$^+$*", "*GatorTron-Large\* + LoRA$^{++}$*", and "*GatorTron-Base*"—demonstrated superior performance compared to SOTA methods. Among these, "*GatorTron-Base + hard-prompting*" achieved the highest performance, surpassing the SOTA model's score of 85.70% by 3.74%. This significant improvement represents a substantial contribution to advancing current research in CTRE systems. This result is beneficial for downstream tasks that build upon CTRE systems, such as EHRs-based data analysis using patients' longitudinal medical histories.

All the model variants that surpassed SOTA performance under fully supervised settings are derived from GatorTron. A key factor contributing to this performance is GatorTron's pretraining on an extensive corpus of over 90 billion words, including more than 82 billion words of de-identified clinical text from the University of Florida Health system [21]. This pretraining aligns with the clinical nature of the i2b2 dataset (derived from Partners HealthCare and the Beth Israel Deaconess Medical Center, MIMIC II Database) [38]. Notably, the i2b2 dataset was not used in GatorTron's pretraining, ensuring the authenticity of its superior performance. Another factor is the suitability of encoder-based models for classification tasks, as opposed to decoder-based models, which are better suited for text generation tasks. Additionally, both "*GatorTron-Base + hard-prompting*" and "*GatorTron-Base*" involve full fine-tuning of the GatorTron-Base model. The performance of "*GatorTron-Base + hard-prompting*," which achieved a 3.59% improvement over "*GatorTron-Base*," demonstrates the efficacy of the proposed fine-tuning strategies.

However, while "*GatorTron-Base + hard-prompting*" yielded the best results, it requires substantial expertise to construct effective hard prompt templates. The performance can vary significantly depending on the specific hard prompt used. When applying the "Model + hard-prompting" method to other CTRE cohorts or even to other relation extraction tasks, it is necessary to redesign the hard prompt in accordance with the characteristics of the new task. Similarly, "*GatorTron-Large\* + LoRA$^+$*" and "*GatorTron-Large\* + LoRA$^{++}$*" also demand a certain level of expertise in prompt engineering to achieve optimal performance. In cases where expertise in prompt engineering or domain knowledge is lacking, models such as "*GatorTron-Base*", which performed slightly higher than SOTA method, may offer greater generalizability and ease of application across different tasks.

Transformer Decoder-based models, such as "LLaMA3-8B/LLaMA3-8B-Instruct" and "MeLLaMA-13B/MeLLaMA-13B-chat," outperformed Transformer Encoder-based models with frozen parameters in our study under the fully supervised setting. This can be attributed to the substantial number of parameters in LLaMA3-8B and MeLLaMA-13B. Even when only certain layers were fine-tuned, the large number of trainable parameters contributed significantly to their strong performance. In contrast, GatorTron-Base, being a lightweight model, did not benefit as much from the incorporation of LoRA, as this method fine-tunes only a small subset of parameters, limiting its ability to adapt effectively to the task. Similarly, with soft-prompting, the fine-tuning capacity was constrained by the fixed prompt vector lengths of 32 or 64. This limitation persisted regardless of whether soft-prompting was applied to GatorTron-Base or GatorTron-Large, resulting in a restricted ability to optimize the model for the specific task.

In few-shot scenarios, the results demonstrate a notable increase in performance from the 1-shot to 4-shot settings, but no significant improvement beyond this point, as performance plateaued between the 4-shot, 8-shot, and 16-shot settings. These findings suggest that while adding a small number of examples can enhance the model's performance, increasing the number of instances beyond a certain threshold yields diminishing returns. Additionally, in few-shot scenarios, Transformer Encoder-based models with frozen parameters outperformed Transformer Decoder-based models with frozen parameters. In other words, models with fewer trainable parameters, including those utilizing soft-prompting techniques and LoRA approaches, are particularly effective when limited training data is available. This can be attributed to the fact that with limited samples, only a small number of parameters can be effectively fine-tuned, and the frozen Encoder-based models are better suited to adapt to such constraints.

There are certain limitations to our study. First, we conducted our experiments solely on the i2b2 CTRE task, which limits the validation of the generalizability of our methods. In future work, we plan to apply our approach to other CTRE tasks, and potentially extend it to other relation extraction tasks. Additionally, we employed lightweight versions of LLaMA3 and MeLLaMA-13B in this study. Exploring larger model variants in future research could provide deeper insights and potentially lead to further performance improvements.

This study explores various fine-tuning strategies applied to four LLMs for CTRE in both few-shot and fully supervised settings. The findings highlight the importance of selecting appropriate models and fine-tuning strategies based on the specific task and data availability to achieve optimal performance. Future work will investigate larger model variants and extend these methods to additional CTRE tasks to enhance generalizability.

**AUTHORS' CONTRIBUTIONS**

CT supervised the study, CT, and JH contributed to the study conception and design, JH, LB, HL, JL, ZS, and EY contributed to data acquisition, processing, and model development, JH was responsible for drafting of the manuscript, CT, DZ, LB, and HL were responsible for critical revision of the manuscript. All authors revised and approved the final manuscript.

## ACKNOWLEDGMENTS

We extend our gratitude to ChatGPT, developed by OpenAI, for providing valuable assistance in refining the language and clarity of this manuscript. Its support in drafting and editing has contributed significantly to the presentation of our work.

## COMPETING INTERESTS

The authors declare no competing interests.

## FUNDING

This research was supported by the National Library of Medicine of the National Institutes of Health under Award Number R01LM014249, National Institute on Aging Awards Numbers R01AG083039 and U01AG070112-02S1, the American Heart Association, Award Number 19GPSGC35180031, University of Illinois at Urbana-Champaign, Award Number 1U01AT012871-01, and, Mayo Clinic Jacksonville, Award Number 7R01AG084236-02.

## FIGURES

**Figure 1.** The architectures of LLMs.
**Figure 2.** The principle of primary fine-tuning strategies for LLMs.
**Figure 3.** The principle of four primary fine-tuning strategies for GatorTron-Base.
**Figure 4.** The principle of two primary PEFT strategies for the Transformer Decoder-based models.
**Figure 5.** The performance (micro F1) of various models under the fully supervised setting. *refers to the parameters inside the model kept frozen; [+] denotes LORA applied exclusively to the query and value layers of the Transformer attention mechanism; [++] indicates that LORA adapted to all linear layers.
**Figure 6.** The performance (micro F1) of various models across different few-shot scenarios. *refers to the parameters inside the model kept frozen; [+] denotes LORA applied exclusively to the query and value layers of the Transformer attention mechanism; [++] indicates that LORA adapted to all linear layers.

## REFERENCES

1   Gumiel YB, Silva e Oliveira LE, Claveau V, *et al.* Temporal Relation Extraction in Clinical Texts: A Systematic Review. *ACM Comput Surv*. 2021;54:1–36. doi: 10.1145/3462475

2   Styler WF 4th, Bethard S, Finan S, *et al.* Temporal Annotation in the Clinical Domain.

*Trans Assoc Comput Linguist*. 2014;2:143–54.

3   Nikfarjam A, Emadzadeh E, Gonzalez G. Towards generating a patient's timeline: extracting temporal relationships from clinical notes. *J Biomed Inform*. 2013;46 Suppl:S40–7. doi: 10.1016/j.jbi.2013.11.001

4   Olex AL, McInnes BT. Review of Temporal Reasoning in the Clinical Domain for Timeline Extraction: Where we are and where we need to be. *J Biomed Inform*. 2021;118:103784. doi: 10.1016/j.jbi.2021.103784

5   Leaman R, Khare R, Lu Z. Challenges in clinical natural language processing for automated disorder normalization. *J Biomed Inform*. 2015;57:28–37. doi: 10.1016/j.jbi.2015.07.010

6   Meystre SM, Savova GK, Kipper-Schuler KC, *et al.* Extracting information from textual documents in the electronic health record: a review of recent research. *Yearb Med Inform*. 2008;128–44.

7   Kreimeyer K, Foster M, Pandey A, *et al.* Natural language processing systems for capturing and standardizing unstructured clinical information: A systematic review. *J Biomed Inform*. 2017;73:14–29. doi: 10.1016/j.jbi.2017.07.012

8   Lin C, Dligach D, Miller TA, *et al.* Multilayered temporal modeling for the clinical domain. *J Am Med Inform Assoc*. 2016;23:387–95. doi: 10.1093/jamia/ocv113

9   Allen JF. Maintaining knowledge about temporal intervals. *Commun ACM*. 1983;26:832–43. doi: 10.1145/182.358434

10  Knez T, Žitnik S. Multimodal learning for temporal relation extraction in clinical texts. *J Am Med Inform Assoc*. 2024;31:1380–7. doi: 10.1093/jamia/ocae059

11  Uma K, Francis S, Moens M-F. Masking language model mechanism with event-driven knowledge graphs for temporal relations extraction from clinical narratives. *Complex Networks & Their Applications XII*. Cham: Springer Nature Switzerland 2024:162–74.

12  Schick T, Schütze H. Exploiting Cloze Questions for Few Shot Text Classification and Natural Language Inference. arXiv [cs.CL]. 2020.

13  Schick T, Schmid H, Schütze H. Automatically Identifying Words That Can Serve as Labels for Few-Shot Text Classification. arXiv [cs.CL]. 2020.

14  Zheng C, Huang M. Exploring Prompt-based Few-shot Learning for Grounded Dialog Generation. arXiv [cs.CL]. 2021.

15  Ma R, Zhou X, Gui T, *et al.* Template-free Prompt Tuning for Few-shot NER. arXiv [cs.CL]. 2021.

16  Sainz O, de Lacalle OL, Labaka G, *et al.* Label Verbalization and Entailment for Effective Zero and Few-Shot Relation Extraction. Proceedings of the 2021 Conference on Empirical

Methods in Natural Language Processing. 2021.

17  Peng C, Yang X, Smith KE, *et al.* Model Tuning or Prompt Tuning? A Study of Large Language Models for Clinical Concept and Relation Extraction. *J Biomed Inform*. 2024;104630. doi: 10.1016/j.jbi.2024.104630

18  Hsu E, Roberts K. Leveraging large language models for knowledge-free weak supervision in clinical natural language processing. *Res Sq*. Published Online First: 28 June 2024. doi: 10.21203/rs.3.rs-4559971/v1

19  Xie Q, Chen Q, Chen A, *et al.* Me LLaMA: Foundation Large Language Models for Medical Applications. arXiv [cs.CL]. 2024.

20  Hu Y, Chen Q, Du J, *et al.* Improving large language models for clinical named entity recognition via prompt engineering. *J Am Med Inform Assoc*. Published Online First: 27 January 2024. doi: 10.1093/jamia/ocad259

21  Yang X, Chen A, PourNejatian N, *et al.* A large language model for electronic health records. *NPJ Digit Med*. 2022;5:194. doi: 10.1038/s41746-022-00742-2

22  Introducing Meta Llama 3: The most capable openly available LLM to date. Meta AI. https://ai.meta.com/blog/meta-llama-3/ (accessed 4 October 2024)

23  Xie Q, Chen Q, Chen A, *et al.* Me-LLaMA: Foundation large language models for medical applications. *Res Sq*. Published Online First: 22 May 2024. doi: 10.21203/rs.3.rs-4240043/v1

24  Zhou L, Friedman C, Parsons S, *et al.* System architecture for temporal information extraction, representation and reasoning in clinical narrative reports. *AMIA Annu Symp Proc*. 2005;869–73.

25  Pustejovsky J, Castano J, Ingria R, *et al.* TimeML: Robust specification of event and temporal expressions in text. https://cdn.aaai.org/Symposia/Spring/2003/SS-03-07/SS03-07-005.pdf (accessed 7 October 2024)

26  Caselli T, Morante R. VUACLTL at Sem Eval 2016 Task 12: A CRF Pipeline to Clinical Temp Eval. https://aclanthology.org/S16-1193.pdf (accessed 16 April 2023)

27  Tang B, Wu Y, Jiang M, *et al.* A hybrid system for temporal information extraction from clinical text. *J Am Med Inform Assoc*. 2013;20:828–35. doi: 10.1136/amiajnl-2013-001635

28  Roberts K, Rink B, Harabagiu SM. A flexible framework for recognizing events, temporal expressions, and temporal relations in clinical text. *J Am Med Inform Assoc*. 2013;20:867–75. doi: 10.1136/amiajnl-2013-001619

29  Tourille J, Ferret O, Névéol A, *et al.* LIMSI-COT at SemEval-2016 Task 12: Temporal relation identification using a pipeline of classifiers. *Proceedings of the 10th International Workshop on Semantic Evaluation (SemEval-2016)*. Stroudsburg, PA, USA: Association for Computational Linguistics 2016.

30    Cohan A, Meurer K, Goharian N. GUIR at SemEval-2016 task 12: Temporal information processing for clinical narratives. *Proceedings of the 10th International Workshop on Semantic Evaluation (SemEval-2016)*. Stroudsburg, PA, USA: Association for Computational Linguistics 2016.

31    Li Z, Li C, Long Y, *et al.* A system for automatically extracting clinical events with temporal information. *BMC Med Inform Decis Mak*. 2020;20:198. doi: 10.1186/s12911-020-01208-9

32    Dligach D, Miller T, Lin C, *et al.* Neural Temporal Relation Extraction. *Proceedings of the 15th Conference of the European Chapter of the Association for Computational Linguistics: Volume 2, Short Papers*. Stroudsburg, PA, USA: Association for Computational Linguistics 2017.

33    Wang Z, Ive J, Velupillai S, *et al.* Is artificial data useful for biomedical Natural Language Processing algorithms? arXiv [cs.CL]. 2019.

34    Lin C, Miller T, Dligach D, *et al.* A BERT-based one-pass multi-task model for clinical temporal relation extraction. *Proceedings of the 19th SIGBioMed Workshop on Biomedical Language Processing*. Stroudsburg, PA, USA: Association for Computational Linguistics 2020.

35    Chen T, Wu M, Li H. A general approach for improving deep learning-based medical relation extraction using a pre-trained model and fine-tuning. *Database* . 2019;2019. doi: 10.1093/database/baz116

36    Haq HU, Kocaman V, Talby D. Deeper clinical document understanding using Relation Extraction. arXiv [cs.CL]. 2021.

37    Dagdelen J, Dunn A, Lee S, *et al.* Structured information extraction from scientific text with large language models. *Nat Commun*. 2024;15:1418. doi: 10.1038/s41467-024-45563-x

38    i2b2: Informatics for Integrating Biology & the Bedside. https://www.i2b2.org/NLP/TemporalRelations/ (accessed 26 September 2024)

39    Hu EJ, Shen Y, Wallis P, *et al.* LoRA: Low-Rank Adaptation of large language models. arXiv [cs.CL]. 2021.

40    Quantization. https://huggingface.co/docs/transformers/en/main_classes/quantization (accessed 6 November 2024)

41    Dettmers T, Pagnoni A, Holtzman A, *et al.* QLoRA: Efficient Finetuning of Quantized LLMs. arXiv [cs.LG]. 2023.

42    He J, Li F, Hu X, *et al.* Chemical-protein relation extraction with pre-trained prompt tuning. *IEEE Int Conf Healthc Inform*. 2022;2022:608–9. doi: 10.1109/ichi54592.2022.00120

43  He J, Li F, Li J, *et al.* Prompt Tuning in Biomedical Relation Extraction. *Int J Healthc Inf Syst Inform*. 2024;8:206–24. doi: 10.1007/s41666-024-00162-9